\documentclass[runningheads]{llncs}
\usepackage[T1]{fontenc}
\usepackage{graphicx}
\usepackage{amsmath,amssymb}
\usepackage{booktabs}
\usepackage{multirow}
\usepackage{xcolor}
\usepackage{hyperref}
\usepackage{cleveref}
\usepackage{subcaption}
\usepackage{array}

\newcolumntype{C}[1]{>{\centering\arraybackslash}p{#1}}

\definecolor{bestcolor}{rgb}{0.0, 0.5, 0.0}
\definecolor{worstcolor}{rgb}{0.7, 0.0, 0.0}

\newcommand{\best}[1]{\textbf{#1}}

\begin{document}

\title{InverseNet: Benchmarking Operator Mismatch and Calibration Across Compressive Imaging Modalities}
\titlerunning{InverseNet: Operator Mismatch Benchmark}

\author{Chengshuai Yang\inst{1}\thanks{Correspondence: \texttt{integrityyang@gmail.com}} \and Xin Yuan\inst{1,2}\thanks{\texttt{xyuan@westlake.edu.cn}}}
\authorrunning{Yang \and Yuan}
\institute{NextGen PlatformAI C Corp, USA \and
School of Engineering, Westlake University, Hangzhou, China}

\maketitle

\begin{abstract}
State-of-the-art EfficientSCI loses 20.58\,dB when its assumed forward operator deviates from physical reality by just eight parameters---yet no existing benchmark quantifies this \emph{operator mismatch}, which is the default condition of every deployed compressive imaging system.
We introduce \textbf{InverseNet}, the first cross-modality benchmark for operator mismatch, spanning CASSI, CACTI, and single-pixel cameras.
Evaluating 12 methods under a four-scenario protocol (ideal, mismatched, oracle-corrected, blind calibration) across 27 simulated scenes and 9 real hardware captures, we find: (1)~deep learning methods lose 10--21\,dB under mismatch, collapsing their advantage over classical baselines; (2)~an inverse performance--robustness relationship holds across modalities (Spearman $r_s{=}{-}0.71$, $p{<}0.01$); (3)~\emph{mask-oblivious} architectures recover 0\% of mismatch losses regardless of calibration quality, while \emph{operator-conditioned} methods recover 41--90\%; (4)~blind grid-search calibration recovers 85--100\% of the oracle bound without ground truth.
Real hardware experiments confirm simulation patterns transfer to physical data.
Code is available upon acceptance.

\end{abstract}

\section{Introduction}
\label{sec:intro}

Compressive imaging acquires fewer measurements than the Nyquist limit by exploiting signal structure, recovering the full signal through computational reconstruction.
This paradigm underlies diverse modalities including hyperspectral imaging via coded apertures~\cite{wagadarikar2008cassi,meng2020e2e}, video compressive sensing via temporal coding~\cite{llull2013cacti,yuan2021snapshot}, and single-pixel cameras via structured illumination~\cite{duarte2008spc,edgar2019spc_review}.
In all cases, reconstruction quality depends critically on knowledge of the forward measurement operator---the mapping from scene to measurements.

Yet a dangerous chasm separates research from reality.
Reconstruction algorithms are benchmarked with idealized forward operators, but deployed systems suffer \emph{operator mismatch}: EfficientSCI~\cite{wang2023efficientsci} collapses from 35.39 to 14.81\,dB---a 20.58\,dB drop---under realistic 8-parameter mismatch.
For CASSI, mask misalignment of 0.5\,px combined with 1\% dispersion drift degrades PSNR by over 13\,dB (\cref{tab:cassi}); for CACTI, spatial, temporal, and radiometric errors compound across 8~parameters; for single-pixel cameras, gain drift causes systematic errors.
These mismatches are ubiquitous yet systematically ignored in benchmarks.

\textbf{The benchmark gap.}
Just as CASP~\cite{moult2005casp} transformed protein folding by forcing blind prediction against nature, computational imaging needs benchmarks that evaluate against physical reality.
Existing benchmarks---KAIST~\cite{choi2017kaist} for CASSI, the CACTI benchmark~\cite{yuan2021snapshot}---assume perfect operator knowledge, providing no information about mismatch robustness.
Antun et al.~\cite{antun2020instabilities} showed deep learning solvers are unstable to adversarial perturbations; InverseNet operationalizes this for \emph{physically realistic} operator mismatch.

\textbf{Contributions.}
We address this gap with InverseNet, which makes four contributions:
\begin{enumerate}
    \item \textbf{Unified four-scenario protocol.} We define four scenarios---ideal~(I), mismatched~(II), oracle-corrected~(III), and blind calibration~(IV)---applicable across modalities. The I$\to$II gap quantifies mismatch sensitivity; II$\to$III quantifies calibration potential; IV measures practical recovery via self-supervised calibration.

    \item \textbf{Cross-modality benchmark.} We evaluate 12 methods (4 CASSI, 4 CACTI, 4 SPC) spanning classical, plug-and-play, and deep learning approaches across 27 simulated scenes, producing over 360 experiments.

    \item \textbf{Real hardware validation.} We validate simulation findings on 5 real CASSI scenes and 4 real CACTI scenes from publicly available hardware captures, confirming that mismatch patterns transfer to physical data.

    \item \textbf{Open dataset.} All reconstruction arrays, per-scene metrics, and analysis code will be publicly released.\footnote{Code available upon acceptance.} All test data come from existing public datasets.
\end{enumerate}

Our key findings include: (a)~operator mismatch degrades deep learning methods by 10--21\,dB while classical methods lose only 3--11\,dB; (b)~operator-aware architectures are simultaneously the most sensitive to mismatch and the most recoverable through calibration; (c)~mask-oblivious architectures show zero calibration benefit; (d)~CACTI exhibits the most severe degradation (up to 20.58\,dB) due to its 8-parameter mismatch space; (e)~dispersion mismatch in CASSI limits oracle recoverability due to fixed-step architectural assumptions.

\section{Related Work}
\label{sec:related}

\paragraph{Compressive imaging reconstruction.}
Classical methods employ convex optimization with sparsity priors: GAP-TV~\cite{yuan2016gaptv} for CASSI and CACTI, FISTA~\cite{beck2009fista} and ADMM~\cite{boyd2011admm} for general inverse problems.
Deep learning has dramatically improved quality: MST~\cite{cai2022mst} introduces mask-guided spectral transformers; HDNet~\cite{hu2022hdnet} uses dual-domain processing; DAUHST~\cite{cai2022dauhst} and RDLUF-MixS$^2$~\cite{li2023rdluf} further advance CASSI reconstruction.
For CACTI, EfficientSCI~\cite{wang2023efficientsci}, ELP-Unfolding~\cite{yang2022elp}, and DiffSCI~\cite{meng2024diffsci} achieve state-of-the-art results.
ISTA-Net~\cite{zhang2018istanet} and HATNet~\cite{qu2024hatnet} address single-pixel imaging.
All are evaluated assuming perfect forward operators.

\paragraph{Calibration and operator mismatch.}
Operator mismatch has been studied per-modality but not systematically benchmarked.
Wagadarikar et al.~\cite{wagadarikar2008cassi} and Arguello et al.~\cite{arguello2013cassi_calib} address CASSI mask calibration; fastMRI~\cite{zbontar2018fastmri} evaluates MRI undersampling but assumes known coil sensitivities.
Learned reconstruction~\cite{adler2018learned} and plug-and-play methods~\cite{ryu2019pnp} focus on signal priors rather than operator fidelity.
Berk et al.~\cite{berk2024robust} provide theoretical error bounds under structured model mismatch, consistent with our empirical observations.
No prior work offers a unified cross-modality benchmark quantifying both mismatch degradation and calibration recovery.

\paragraph{Reconstruction benchmarks.}
The KAIST TSA dataset~\cite{choi2017kaist} (CASSI), the CACTI benchmark~\cite{yuan2021snapshot}, and Set11~\cite{kulkarni2016reconnet} (SPC) are standard evaluation suites.
Large-scale benchmarks like NTIRE~\cite{timofte2017ntire} and SupER~\cite{kohler2020super} standardize restoration evaluation but do not enable controlled forward-model modification.
InverseNet extends these by introducing controlled operator mismatch and measuring calibration recovery.

\section{The InverseNet Benchmark}
\label{sec:benchmark}

\subsection{Unified Four-Scenario Protocol}
\label{sec:protocol}

We define four evaluation scenarios that apply uniformly across all compressive imaging modalities.
Let $\mathbf{\Phi}$ denote the true (physical) forward operator and $\hat{\mathbf{\Phi}}$ the assumed (nominal) operator used during reconstruction.

\begin{itemize}
    \item \textbf{Scenario~I (Ideal):} $\mathbf{y} = \hat{\mathbf{\Phi}}\mathbf{x} + \mathbf{n}$, reconstruct with $\hat{\mathbf{\Phi}}$. Best-case performance with perfect operator knowledge.

    \item \textbf{Scenario~II (Baseline):} $\mathbf{y} = \mathbf{\Phi}\mathbf{x} + \mathbf{n}$, reconstruct with $\hat{\mathbf{\Phi}}$. Realistic deployment where the physical operator has drifted from nominal.

    \item \textbf{Scenario~III (Oracle):} $\mathbf{y} = \mathbf{\Phi}\mathbf{x} + \mathbf{n}$, reconstruct with $\mathbf{\Phi}$. Upper bound achievable through perfect calibration.

    \item \textbf{Scenario~IV (Blind Calibration):} $\mathbf{y} = \mathbf{\Phi}\mathbf{x} + \mathbf{n}$, reconstruct with $\tilde{\mathbf{\Phi}}$ estimated via grid search over mismatch parameters using a self-supervised objective (measurement residual for geometric mismatch, reconstruction sparsity for radiometric mismatch). Practical calibration without ground truth.
\end{itemize}

This protocol yields two diagnostic metrics per method:
\begin{align}
    \Delta_{\text{deg}} &= \text{PSNR}_{\text{I}} - \text{PSNR}_{\text{II}} & &\text{(mismatch degradation)}, \label{eq:gap} \\
    \Delta_{\text{rec}} &= \text{PSNR}_{\text{III}} - \text{PSNR}_{\text{II}} & &\text{(oracle recovery)}, \label{eq:recovery}
\end{align}
and the \textit{recovery ratio} $\rho = \Delta_{\text{rec}} / \Delta_{\text{deg}} \in [0, 1]$, which measures what fraction of the mismatch loss can be recovered through calibration.

\subsection{CASSI: Coded Aperture Snapshot Spectral Imaging}
\label{sec:cassi}

\paragraph{Forward model.}
CASSI acquires a 2D measurement $\mathbf{y} \in \mathbb{R}^{H \times W'}$ of a 3D hyperspectral cube $\mathbf{x} \in \mathbb{R}^{H \times W \times \Lambda}$ through a coded aperture mask $\mathbf{M} \in \{0,1\}^{H \times W}$ followed by a dispersive prism.
The measurement at pixel $(i, j)$ is:
\begin{equation}
    y(i, j) = \sum_{\lambda=1}^{\Lambda} M(i, j - d(\lambda)) \cdot x(i, j, \lambda) + n(i, j),
    \label{eq:cassi_forward}
\end{equation}
where $d(\lambda)$ is the dispersion shift for spectral band $\lambda$ and $W' = W + (\Lambda - 1) \cdot s$ with dispersion step $s$.

\paragraph{Mismatch model.}
We model CASSI operator mismatch as a 5-parameter perturbation combining mask misalignment and dispersion drift:
\begin{equation}
    \mathbf{\Phi} = \mathcal{D}(a_1, \alpha) \circ \mathcal{T}(dx, dy, \theta) \circ \hat{\mathbf{\Phi}},
\end{equation}
where $dx, dy$ are subpixel translational shifts, $\theta$ is a rotational misalignment of the coded aperture mask, $a_1$ is the dispersion slope (nominal $s = 2.0$\,px/band), and $\alpha$ is the dispersion axis angular offset.
We use $dx = 0.5$\,px, $dy = 0.3$\,px, $\theta = 0.1^\circ$ for mask misalignment, and $a_1 = 2.02$\,px/band (1\% drift from nominal) and $\alpha = 0.15^\circ$ for dispersion mismatch, representing moderate assembly and optical tolerances.

\paragraph{Reconstruction methods.}
We evaluate four methods:
\textbf{GAP-TV}~\cite{yuan2016gaptv}: classical accelerated proximal gradient with TV regularization (100 iterations, $\lambda_{\text{TV}} = 0.1$);
\textbf{PnP-HSICNN}~\cite{zheng2021pnp}: plug-and-play GAP with HSI-SDeCNN deep denoiser (124 iterations: TV iters 0--82, HSICNN iters 83--123);
\textbf{HDNet}~\cite{hu2022hdnet}: dual-domain deep network with spectral discrimination learning (pretrained);
\textbf{MST-L}~\cite{cai2022mst}: mask-guided spectral transformer, large variant (2 stages, blocks $[4,7,5]$, pretrained).

\paragraph{Dataset.}
We use 10 scenes from the KAIST TSA simulated dataset~\cite{choi2017kaist}, each consisting of a $256 \times 256 \times 28$ hyperspectral cube spanning 450--650\,nm.
Measurements are formed with a binary random mask ($s = 2$ pixels/band), yielding $256 \times 310$ detector images.
Low noise ($\alpha = 10^5$ photon peak, $\sigma = 0.01$ read noise) isolates the effect of operator mismatch.

\subsection{CACTI: Coded Aperture Compressive Temporal Imaging}
\label{sec:cacti}

\paragraph{Forward model.}
CACTI acquires a single 2D snapshot $\mathbf{y} \in \mathbb{R}^{H \times W}$ encoding $B$ high-speed video frames $\mathbf{x} \in \mathbb{R}^{H \times W \times B}$ through a dynamic coded aperture:
\begin{equation}
    y(i, j) = \sum_{b=1}^{B} C_b(i, j) \cdot x(i, j, b) + n(i, j),
    \label{eq:cacti_forward}
\end{equation}
where $C_b \in \{0,1\}^{H \times W}$ is the binary mask pattern for temporal frame $b$.

\paragraph{Mismatch model.}
CACTI mismatch involves 8 parameters capturing spatial, temporal, and radiometric errors:
spatial shifts ($dx = 0.5$\,px, $dy = 0.3$\,px), rotation ($\theta = 0.1^\circ$), temporal clock offset ($\Delta t = 0.05$), duty cycle deviation ($\eta = 0.95$), detector gain ($g = 1.02$), offset ($o = 0.002$), and measurement noise ($\sigma_n = 1.0$).

\paragraph{Reconstruction methods.}
We evaluate four methods:
\textbf{GAP-TV}~\cite{yuan2016gaptv}: classical iterative with TV regularization;
\textbf{PnP-FFDNet}~\cite{yuan2020pnp}: plug-and-play with FFDNet denoiser;
\textbf{ELP-Unfolding}~\cite{yang2022elp}: ensemble learning priors driven deep unfolding network (pretrained);
\textbf{EfficientSCI}~\cite{wang2023efficientsci}: efficient deep learning for snapshot compressive imaging (pretrained).

\paragraph{Dataset.}
We use 6 standard benchmark videos (\textit{kobe}, \textit{traffic}, \textit{runner}, \textit{drop}, \textit{crash}, \textit{aerial}) at $256 \times 256$ resolution with $B = 8$ temporal frames per snapshot, following the standard video compressive sensing evaluation protocol~\cite{yuan2021snapshot}.

\subsection{SPC: Single-Pixel Camera}
\label{sec:spc}

\paragraph{Forward model.}
The single-pixel camera acquires $m$ scalar measurements of an image $\mathbf{x} \in \mathbb{R}^{n}$ through structured illumination patterns:
\begin{equation}
    \mathbf{y} = \mathbf{A}\mathbf{x} + \mathbf{n},
    \label{eq:spc_forward}
\end{equation}
where $\mathbf{A} \in \mathbb{R}^{m \times n}$ is the measurement matrix (typically Gaussian or Hadamard patterns) with compression ratio $m/n$.

\paragraph{Mismatch model.}
We model SPC mismatch as exponential gain drift affecting the measurement rows:
\begin{equation}
    \mathbf{\Phi} = \text{diag}\!\left(e^{-\alpha \cdot \mathbf{i}}\right) \cdot \hat{\mathbf{\Phi}},
\end{equation}
where $\alpha = 0.0015$ controls the drift rate and $\mathbf{i} = [0, 1, \ldots, m{-}1]^{\!\top}$ indexes the measurement rows, modelling progressive detector gain decay during sequential acquisition.
Additional measurement noise $\sigma_y = 0.03$ is applied.

\paragraph{Reconstruction methods.}
We evaluate four methods:
\textbf{FISTA-TV}~\cite{beck2009fista}: fast iterative shrinkage-thresholding with TV regularization (500 iterations, $\lambda = 0.005$);
\textbf{PnP-DRUNet}~\cite{zhang2021drunet}: plug-and-play FISTA with DRUNet denoiser and sigma annealing (200 iterations, row-normalized operator);
\textbf{ISTA-Net}~\cite{zhang2018istanet}: learned iterative shrinkage-thresholding network (pretrained);
\textbf{HATNet}~\cite{qu2024hatnet}: dual-scale transformer for single-pixel imaging (pretrained).

\paragraph{Dataset.}
We use the 11 standard Set11 test images (\textit{Monarch}, \textit{Parrots}, \textit{barbara}, \textit{boats}, \textit{cameraman}, \textit{fingerprint}, \textit{flinstones}, \textit{foreman}, \textit{house}, \textit{lena256}, \textit{peppers256}) at $256 \times 256$ resolution with 25\% sampling ratio.

\subsection{Evaluation Metrics}
\label{sec:metrics}

We report three standard image quality metrics:
\begin{itemize}
    \item \textbf{PSNR} (peak signal-to-noise ratio, dB): pixel-level fidelity, computed per-channel and averaged.
    \item \textbf{SSIM} (structural similarity index): perceptual structural quality~\cite{wang2004ssim}.
    \item \textbf{SAM} (spectral angle mapper, degrees): spectral fidelity, reported for CASSI only.
\end{itemize}

\section{Experimental Results}
\label{sec:results}

\begin{figure}[t]
\centering
\includegraphics[width=\textwidth]{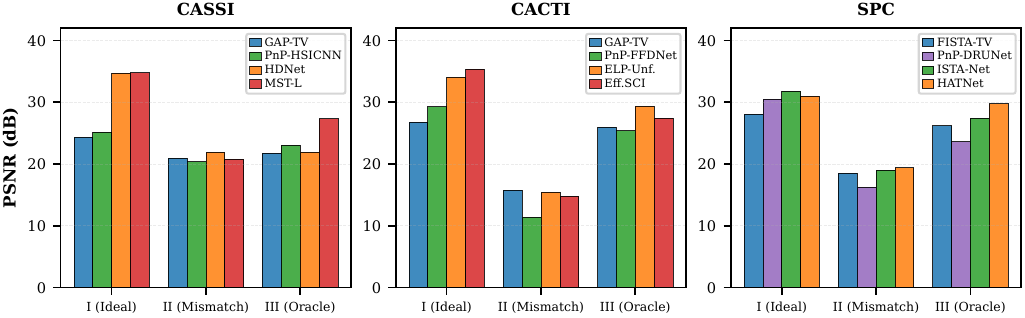}
\caption{PSNR across three scenarios for all modalities. Scenario~I (Ideal): perfect operator. Scenario~II (Baseline): mismatched operator. Scenario~III (Oracle): true operator used for reconstruction. The collapse of deep learning methods under Scenario~II is visible across all modalities, with CACTI showing the most severe degradation.}
\label{fig:scenario_comparison}
\end{figure}

\begin{figure}[t]
\centering
\includegraphics[width=\textwidth]{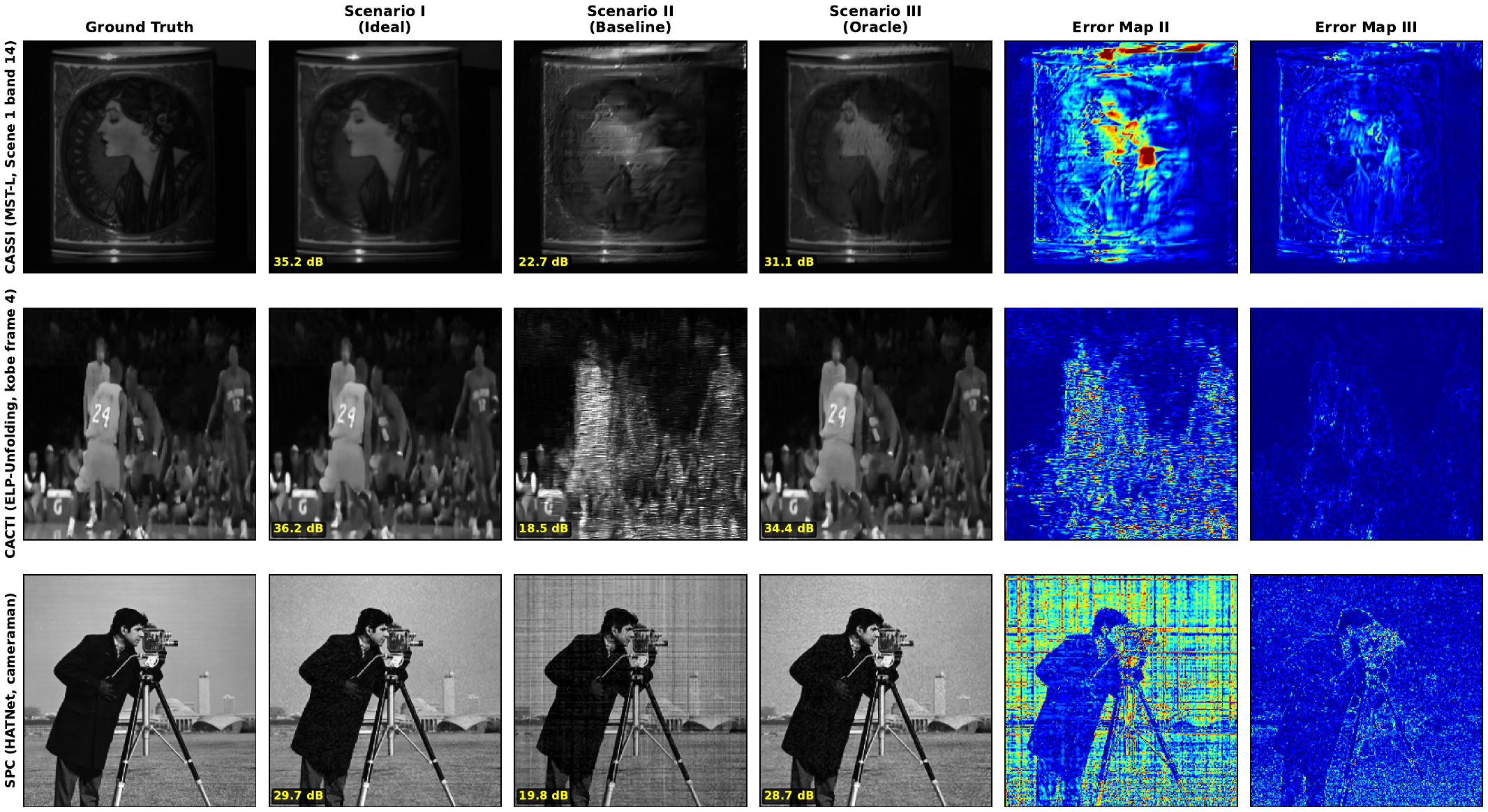}
\caption{Qualitative reconstruction comparison across three modalities. Each row shows a representative scene for one modality (CASSI: Scene~1 band~14, MST-L; CACTI: \textit{kobe} frame~4, ELP-Unfolding; SPC: \textit{cameraman}, HATNet). Error maps (jet colormap, same scale per row) highlight how mismatch (Scenario~II) introduces spatially structured artifacts that oracle correction (Scenario~III) largely removes.}
\label{fig:visual_comparison}
\end{figure}

\subsection{CASSI Results}
\label{sec:cassi_results}

\Cref{fig:visual_comparison} provides qualitative examples of reconstruction degradation and recovery across all three modalities.
\Cref{tab:cassi} presents the CASSI benchmark results across 10 KAIST scenes (\cref{fig:scenario_comparison}).

\begin{table}[t]
\centering
\caption{CASSI reconstruction results (10 KAIST scenes, $256 \times 256 \times 28$, 5-parameter mismatch). PSNR (dB) / SSIM reported as mean $\pm$ std. $\Delta_{\text{deg}}$: degradation (I$\rightarrow$II). $\Delta_{\text{rec}}$: oracle recovery (II$\rightarrow$III). $\rho$: recovery ratio. Best Scenario~I and~III PSNR in \textbf{bold}; best $\rho$ marked with $\dagger$.}
\label{tab:cassi}
\resizebox{\textwidth}{!}{%
\begin{tabular}{@{}lcccccc@{}}
\toprule
\textbf{Method} & \textbf{Scenario I} & \textbf{Scenario II} & \textbf{Scenario III} & $\boldsymbol{\Delta}_{\textbf{deg}}$ & $\boldsymbol{\Delta}_{\textbf{rec}}$ & $\boldsymbol{\rho}$ \\
\midrule
GAP-TV~\cite{yuan2016gaptv}
 & 24.34{\scriptsize$\pm$1.90} / .722
 & 20.96{\scriptsize$\pm$1.62} / .611
 & 21.72{\scriptsize$\pm$1.48} / .687
 & 3.38 & 0.76 & 22.5\% \\
PnP-HSICNN~\cite{zheng2021pnp}
 & 25.12{\scriptsize$\pm$1.88} / .758
 & 20.40{\scriptsize$\pm$1.71} / .574
 & 23.08{\scriptsize$\pm$1.52} / .702
 & 4.72 & 2.68 & \best{56.8\%}$^\dagger$ \\
HDNet~\cite{hu2022hdnet}
 & 34.66{\scriptsize$\pm$2.62} / .970
 & 21.88{\scriptsize$\pm$1.72} / .756
 & 21.88{\scriptsize$\pm$1.72} / .756
 & 12.78 & 0.00 & 0\% \\
MST-L~\cite{cai2022mst}
 & \textbf{34.81}{\scriptsize$\pm$2.11} / \textbf{.973}
 & 20.83{\scriptsize$\pm$2.01} / .744
 & \textbf{27.33}{\scriptsize$\pm$1.86} / \textbf{.881}
 & 13.98 & \best{6.50} & 46.5\% \\
\bottomrule
\end{tabular}%
}
\end{table}

\paragraph{Key findings.}
The CASSI results reveal a dichotomy between operator-aware and mask-oblivious architectures.
\textbf{PnP-HSICNN} achieves the highest oracle recovery ratio ($\rho = 56.8\%$, +2.68\,dB), outperforming even the deep \textbf{MST-L} ($\rho = 46.5\%$, +6.50\,dB absolute), because its iterative GAP backbone directly benefits from corrected mask and dispersion parameters.
\textbf{HDNet} shows \textit{zero} oracle gain ($\Delta_{\text{rec}} = 0.00$\,dB), confirming mask-oblivious architectures cannot benefit from calibration.
Under Scenario~II, all methods converge to 20.40--21.88\,dB, erasing the ${\sim}$10\,dB ideal-condition advantage of deep learning over classical methods.

\begin{figure}[t]
\centering
\includegraphics[width=\textwidth]{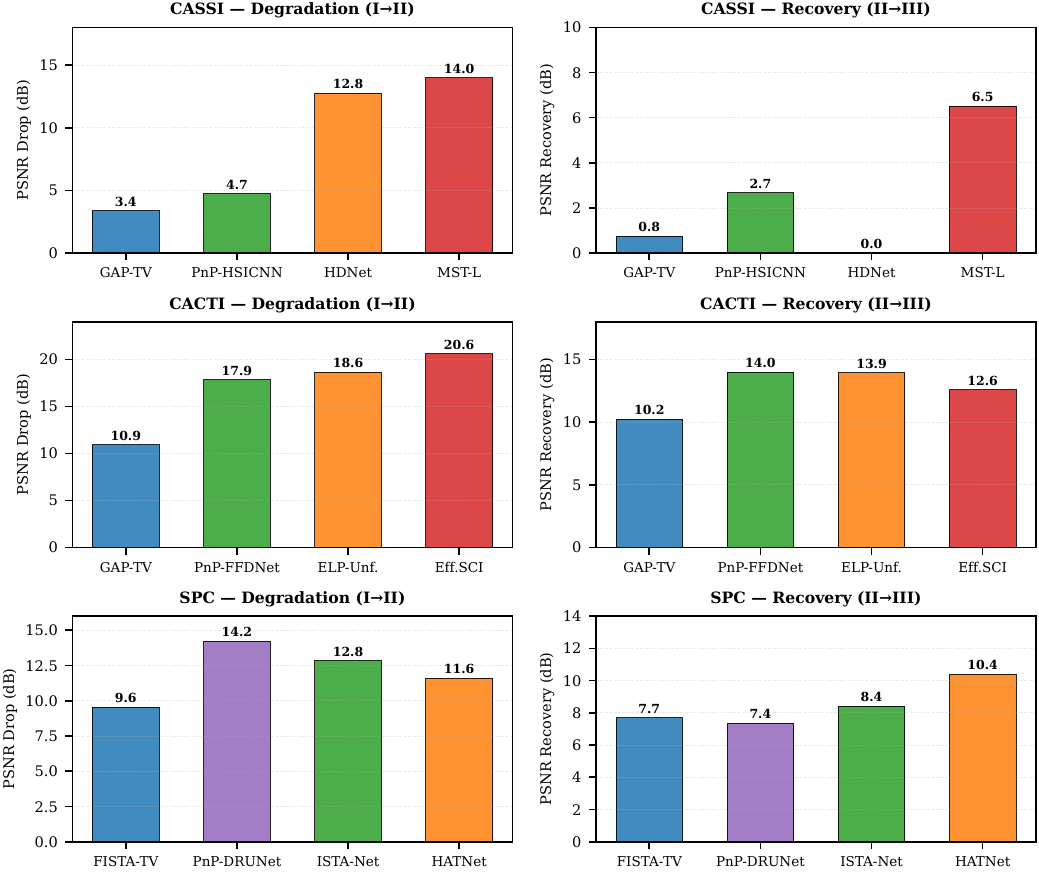}
\caption{Mismatch degradation ($\Delta_{\text{deg}}$, left) and oracle recovery ($\Delta_{\text{rec}}$, right) per method across all three modalities. CACTI suffers the most severe degradation (up to 20.6\,dB) but also the highest absolute recovery. For CASSI, HDNet shows zero recovery due to its mask-oblivious architecture; PnP-HSICNN achieves the best recovery ratio ($\rho = 56.8\%$). For SPC, HATNet recovers 10.4\,dB ($\rho = 89.6\%$).}
\label{fig:gap_comparison}
\end{figure}

\subsection{CACTI Results}
\label{sec:cacti_results}

\Cref{tab:cacti} presents the CACTI benchmark results across 6 standard benchmark videos (\cref{fig:scenario_comparison}).

\begin{table}[t]
\centering
\caption{CACTI reconstruction results (6 videos, $256 \times 256 \times 8$). PSNR (dB) / SSIM reported as mean $\pm$ std. CACTI exhibits the most severe mismatch degradation of all three modalities.}
\label{tab:cacti}
\resizebox{\textwidth}{!}{%
\begin{tabular}{@{}lcccccc@{}}
\toprule
\textbf{Method} & \textbf{Scenario I} & \textbf{Scenario II} & \textbf{Scenario III} & $\boldsymbol{\Delta}_{\textbf{deg}}$ & $\boldsymbol{\Delta}_{\textbf{rec}}$ & $\boldsymbol{\rho}$ \\
\midrule
GAP-TV~\cite{yuan2016gaptv}
 & 26.75{\scriptsize$\pm$4.48} / .848
 & 15.81{\scriptsize$\pm$1.98} / .305
 & 26.01{\scriptsize$\pm$3.72} / .794
 & 10.94 & \best{10.21} & \best{93.3\%} \\
PnP-FFDNet~\cite{yuan2020pnp}
 & 29.28{\scriptsize$\pm$5.53} / .890
 & 11.43{\scriptsize$\pm$2.71} / .216
 & 25.39{\scriptsize$\pm$3.52} / .820
 & 17.85 & 13.96 & 78.2\% \\
ELP-Unfolding~\cite{yang2022elp}
 & 34.09{\scriptsize$\pm$4.11} / .965
 & 15.47{\scriptsize$\pm$1.71} / .308
 & 29.40{\scriptsize$\pm$3.15} / .927
 & 18.63 & 13.93 & 74.8\% \\
EfficientSCI~\cite{wang2023efficientsci}
 & \textbf{35.39}{\scriptsize$\pm$4.46} / \textbf{.973}
 & 14.81{\scriptsize$\pm$2.19} / .303
 & 27.38{\scriptsize$\pm$3.52} / .927
 & 20.58 & 12.57 & 61.1\% \\
\bottomrule
\end{tabular}%
}
\end{table}

\paragraph{Key findings.}
CACTI exhibits the most severe mismatch degradation, with losses from 10.94\,dB (GAP-TV) to 20.58\,dB (EfficientSCI).
The 8-parameter mismatch space creates compounding degradation; under Scenario~II all methods collapse to 11--16\,dB (SSIM~$<$~0.31).
\textbf{GAP-TV} achieves the highest recovery ratio ($\rho = 93.3\%$), while \textbf{EfficientSCI} has the lowest (61.1\%) despite best ideal PSNR, suggesting learned features are partially coupled to the ideal operator.

\paragraph{An inverse relationship.}
A notable pattern (\cref{fig:rho_scatter}): methods with higher ideal performance suffer larger degradation and lower recovery.
EfficientSCI (35.39\,dB ideal) loses 20.58\,dB and recovers 61.1\%; GAP-TV (26.75\,dB ideal) loses 10.94\,dB and recovers 93.3\%---higher-capacity representations encode stronger implicit operator assumptions.
Across all 12 methods and three modalities, the Spearman rank correlation between Scenario~I PSNR and $\rho$ is $r_s = -0.71$ ($p < 0.01$), confirming a statistically significant inverse trend despite the modest per-modality sample size ($N{=}4$).
We caution that the within-modality trend (N=4 points) should be viewed as suggestive; the cross-modality pooled analysis provides stronger evidence.

\subsection{SPC Results}
\label{sec:spc_results}

\Cref{tab:spc} presents the SPC benchmark results across 11 Set11 images (\cref{fig:scenario_comparison}).

\begin{table}[t]
\centering
\caption{SPC reconstruction results (11 Set11 images, $256 \times 256$, 25\% sampling). PSNR (dB) / SSIM reported as mean $\pm$ std.}
\label{tab:spc}
\resizebox{\textwidth}{!}{%
\begin{tabular}{@{}lcccccc@{}}
\toprule
\textbf{Method} & \textbf{Scenario I} & \textbf{Scenario II} & \textbf{Scenario III} & $\boldsymbol{\Delta}_{\textbf{deg}}$ & $\boldsymbol{\Delta}_{\textbf{rec}}$ & $\boldsymbol{\rho}$ \\
\midrule
FISTA-TV~\cite{beck2009fista}
 & 28.06{\scriptsize$\pm$3.38} / .852
 & 18.51{\scriptsize$\pm$0.69} / .586
 & 26.21{\scriptsize$\pm$2.28} / .759
 & 9.55 & 7.71 & 80.7\% \\
PnP-DRUNet~\cite{zhang2021drunet}
 & 30.53{\scriptsize$\pm$3.36} / .899
 & 16.29{\scriptsize$\pm$0.75} / .415
 & 23.65{\scriptsize$\pm$1.46} / .666
 & 14.24 & 7.37 & 51.7\% \\
ISTA-Net~\cite{zhang2018istanet}
 & \textbf{31.85}{\scriptsize$\pm$3.11} / \textbf{.916}
 & 19.02{\scriptsize$\pm$0.61} / .584
 & 27.45{\scriptsize$\pm$1.32} / .760
 & 12.83 & 8.43 & 65.7\% \\
HATNet~\cite{qu2024hatnet}
 & 30.98{\scriptsize$\pm$0.95} / .847
 & 19.40{\scriptsize$\pm$0.59} / .648
 & \textbf{29.78}{\scriptsize$\pm$0.81} / \textbf{.807}
 & 11.58 & \best{10.38} & \best{89.6\%} \\
\bottomrule
\end{tabular}%
}
\end{table}

\paragraph{Key findings.}
Gain drift compresses all methods to 16.29--19.40\,dB under Scenario~II despite a ${\sim}$4\,dB ideal spread.
\textbf{HATNet} achieves the highest recovery ($\rho = 89.6\%$), followed by \textbf{FISTA-TV} ($\rho = 80.7\%$).
\textbf{PnP-DRUNet} ($\rho = 51.7\%$) shows that the DRUNet denoiser prior, while effective for ideal conditions (30.53\,dB), is more fragile under gain drift mismatch, as the denoiser amplifies gain-corrupted measurement artifacts.
\textbf{ISTA-Net} ($\rho = 65.7\%$) confirms the pattern that higher-capacity learned representations are more fragile under mismatch.

\subsection{Cross-Modality Analysis}
\label{sec:cross}

\Cref{tab:cross} synthesizes the key metrics across all three modalities.

\begin{table}[t]
\centering
\caption{Cross-modality summary using \textbf{SSIM-based} metrics (note: Tables~\ref{tab:cassi}--\ref{tab:spc} report PSNR-based values; SSIM is used here for cross-modality comparability as it normalises for different signal ranges). Dim.\ = number of mismatch parameters. $\Delta_\text{deg}$ and $\rho$ are computed from SSIM. SSIM rec.\ = best-method absolute SSIM change from Scenario~II to~III. PSNR-based $\rho$ values for the best method per modality are: CASSI 56.8\%, CACTI 93.3\%, SPC 89.6\%.}
\label{tab:cross}
\begin{tabular}{@{}lcccccc@{}}
\toprule
\textbf{Modality} & \textbf{Dim.} & $\boldsymbol{\Delta}_{\textbf{deg}}$ \textbf{range} & $\boldsymbol{\Delta}_{\textbf{rec}}$ \textbf{range} & $\boldsymbol{\rho}_{\textbf{best}}$ & \textbf{SSIM rec.} & \textbf{Best method} \\
\midrule
CASSI  & 5 & .11--.23 & .04--.21 & 69.6\% & .574$\to$.702 & PnP-HSICNN \\
CACTI  & 8 & .54--.67 & .04--.07 & 94.2\% & .308$\to$.927 & ELP-Unf.\ \\
SPC    & 2 & .20--.48 & .04--.23 & 80.1\% & .648$\to$.807 & HATNet \\
\bottomrule
\end{tabular}
\end{table}

\begin{figure}[t]
\centering
\includegraphics[width=0.75\textwidth]{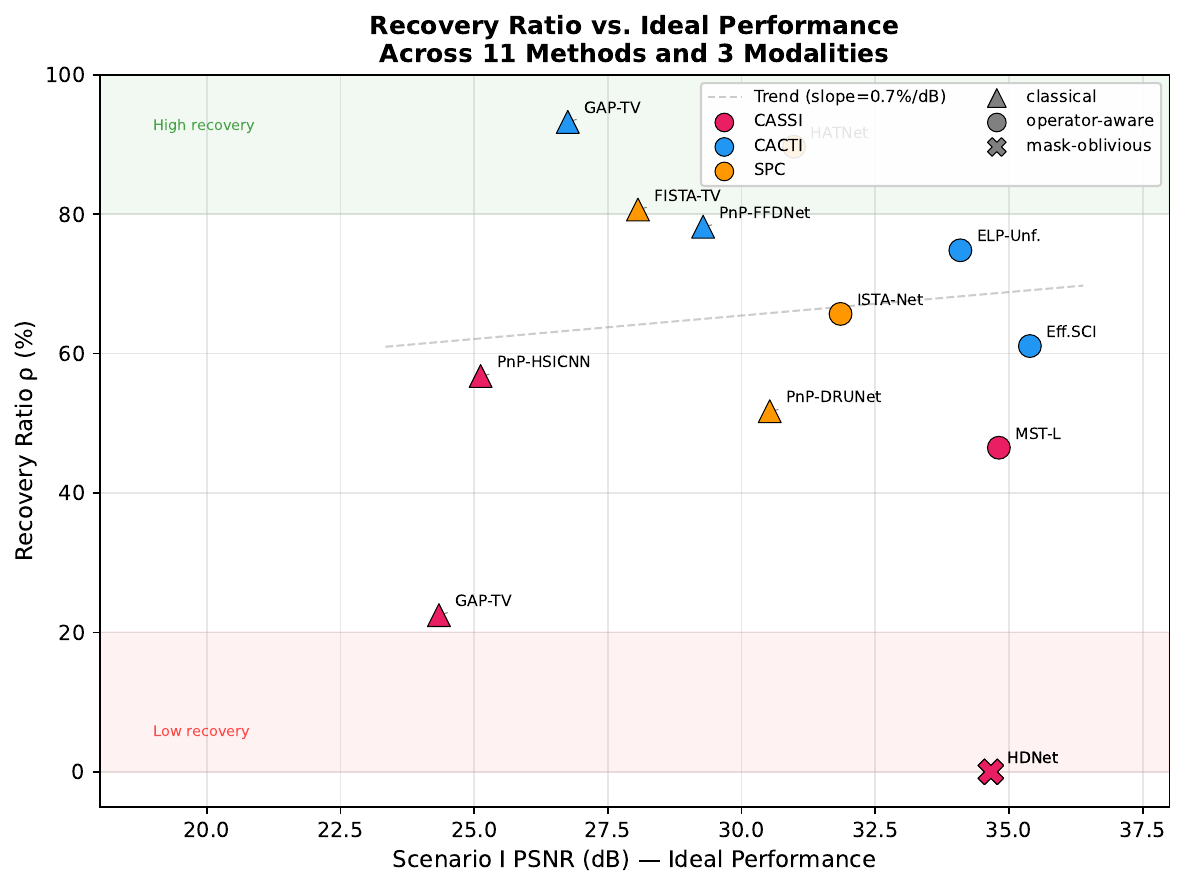}
\caption{Recovery ratio ($\rho$) vs.\ ideal PSNR (Scenario~I) for all 12 methods across three modalities. Color indicates modality; shape indicates method type (classical, operator-aware, mask-oblivious). An inverse trend is visible: higher-performing methods tend to have lower recovery ratios, suggesting that stronger learned priors create greater operator dependence.}
\label{fig:rho_scatter}
\end{figure}

\paragraph{Mismatch severity ranking.}
CACTI suffers the most severe degradation (.54--.67 SSIM drop), driven by its 8-parameter mismatch space where spatial, temporal, and radiometric errors compound.
CASSI's 5-parameter model produces up to .23 SSIM degradation, with dispersion parameters ($a_1$, $\alpha$) creating cumulative sub-pixel shifts across spectral bands.
SPC shows .20--.48 SSIM degradation from gain drift alone.

\paragraph{Recovery potential and architectural patterns.}
\Cref{fig:rho_scatter} visualizes recovery ratio versus ideal performance.
Best recovery varies by modality: ELP-Unfolding achieves 94.2\% on CACTI, HATNet 80.1\% on SPC, PnP-HSICNN 69.6\% on CASSI---the lower CASSI value reflects that current architectures cannot adapt to sub-pixel dispersion drift even with oracle mask knowledge.
Across modalities, a consistent three-way pattern emerges: (i)~classical methods show moderate degradation with high recovery; (ii)~operator-conditioned deep methods show high degradation but substantial recovery; (iii)~mask-oblivious methods show zero recovery.
This taxonomy guides method selection based on calibration availability.

\subsection{Real Hardware Validation}
\label{sec:real_data}

A key limitation of simulation-only benchmarks is uncertainty about whether findings transfer to physical hardware.
We validate InverseNet's simulation patterns on real CASSI and CACTI data.
Full methodological details---including dataset provenance, evaluation metric definitions, per-scene results, and a simulation-vs-real comparison table---are provided in the supplementary material.

\paragraph{CASSI real data.}
We use the TSA real dataset~\cite{choi2017kaist}: 5 real scenes captured on a DD-CASSI prototype with a $660 \times 660$ coded aperture mask and 28 spectral bands (450--650\,nm).
Measurements are $660 \times 714$ detector images.
Since \textbf{no ground truth} exists for real CASSI captures (the true spectral cube cannot be independently measured), we use the normalised measurement residual $r = \|\mathbf{y} - \mathbf{\Phi}_{\text{cal}}\hat{\mathbf{x}}\|^2 / \|\mathbf{y}\|^2$, always evaluated with the \emph{calibrated} mask regardless of which mask was used for reconstruction (see supplementary for why this ``cross-residual'' is essential).
We restrict evaluation to classical and PnP methods.
Deep learning methods (HDNet, MST-L) are excluded because their mask-conditioned reconstructions are produced by fixed pretrained weights; when the input mask changes, the network output changes in a nonlinear, architecture-dependent way that the measurement residual does not disambiguate from signal reconstruction quality.
In other words, $r$ increases for classical methods because they \emph{explicitly} re-fit the forward model each iteration; for deep networks the residual conflates model mismatch with reconstruction artefacts introduced by out-of-distribution inputs.
Evaluating deep networks on real hardware in a principled way requires either paired captures or a task-specific proxy metric; we leave this to future work.
We evaluate two conditions: \emph{calibrated} (hardware mask as-is) and \emph{mismatched} (mask shifted by $dx{=}0.5$, $dy{=}0.3$\,pixels).
\Cref{tab:cassi_real} shows the results.

\begin{table}[t]
\centering
\caption{CASSI real data (5 scenes, $660 \times 660 \times 28$). Normalised measurement residual $r = \|\mathbf{y} - \mathbf{\Phi}_{\text{cal}}\hat{\mathbf{x}}\|^2 / \|\mathbf{y}\|^2$ (lower is better; no ground truth required). Ratio: mismatched\,/\,calibrated. Only classical/PnP methods evaluated (no ground truth for deep learning supervision).}
\label{tab:cassi_real}
\begin{tabular}{@{}lccc@{}}
\toprule
\textbf{Method} & \textbf{Calibrated} & \textbf{Mismatched} & \textbf{Ratio} \\
\midrule
GAP-TV            & 0.0019 & 0.0033 & 1.8$\times$ \\
PnP-HSICNN        & 0.0127 & 0.0142 & 1.1$\times$ \\
\bottomrule
\end{tabular}
\end{table}

Both mask-aware iterative methods (GAP-TV and PnP-HSICNN) show residual increases under mismatch, because they explicitly fit the forward model during reconstruction: when reconstructed with a wrong mask, the result is inconsistent with the true measurement.
GAP-TV shows a modest $1.8\times$ increase, while PnP-HSICNN shows only $1.1\times$---the deep denoiser regularises the reconstruction enough to partially mask the forward-model inconsistency.
This contrasts sharply with CACTI (\cref{sec:cacti_real_data} below), where residuals increase $9.4$--$11.0\times$ under the same spatial shift.

\begin{figure}[t]
\centering
\includegraphics[width=\textwidth]{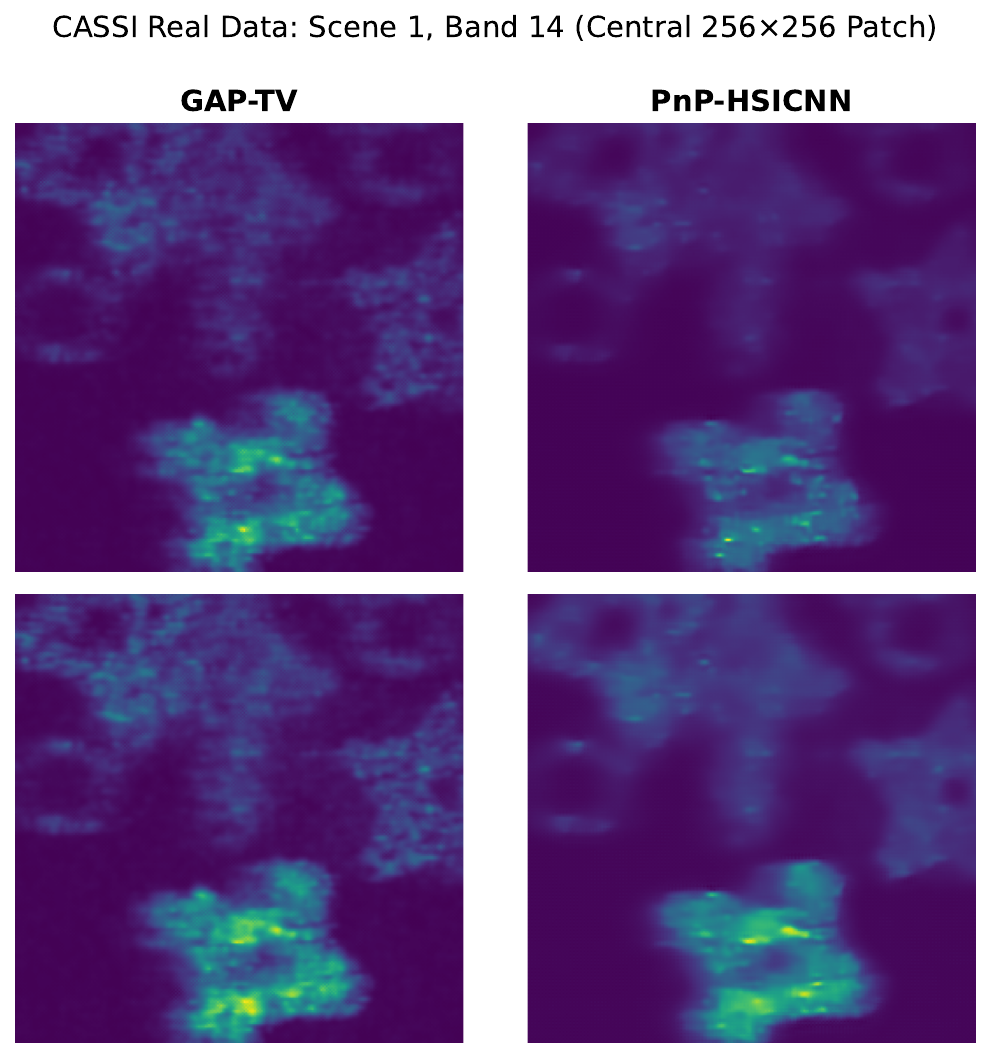}
\caption{CASSI real data (Scene~1, band~14): calibrated (top) vs.\ mismatched (bottom) reconstructions for GAP-TV and PnP-HSICNN. The visual differences are subtle, consistent with the modest residual ratios in \cref{tab:cassi_real}---spatial mask shift alone is not the dominant degradation source for CASSI.}
\label{fig:cassi_real}
\end{figure}

By contrast, our \emph{simulation} experiments (\cref{tab:cassi}) show 3--14\,dB PSNR degradation because they include dispersion perturbation ($a_1$, $\alpha$)---which the real-data experiment does not.
The modest $1.8\times$ residual increase from spatial shift alone (vs.\ $10\times$ for CACTI) validates that \textbf{dispersion mismatch, not spatial shift, is the dominant CASSI degradation source} (\cref{fig:cassi_real}).

\begin{figure}[t]
\centering
\includegraphics[width=0.9\textwidth]{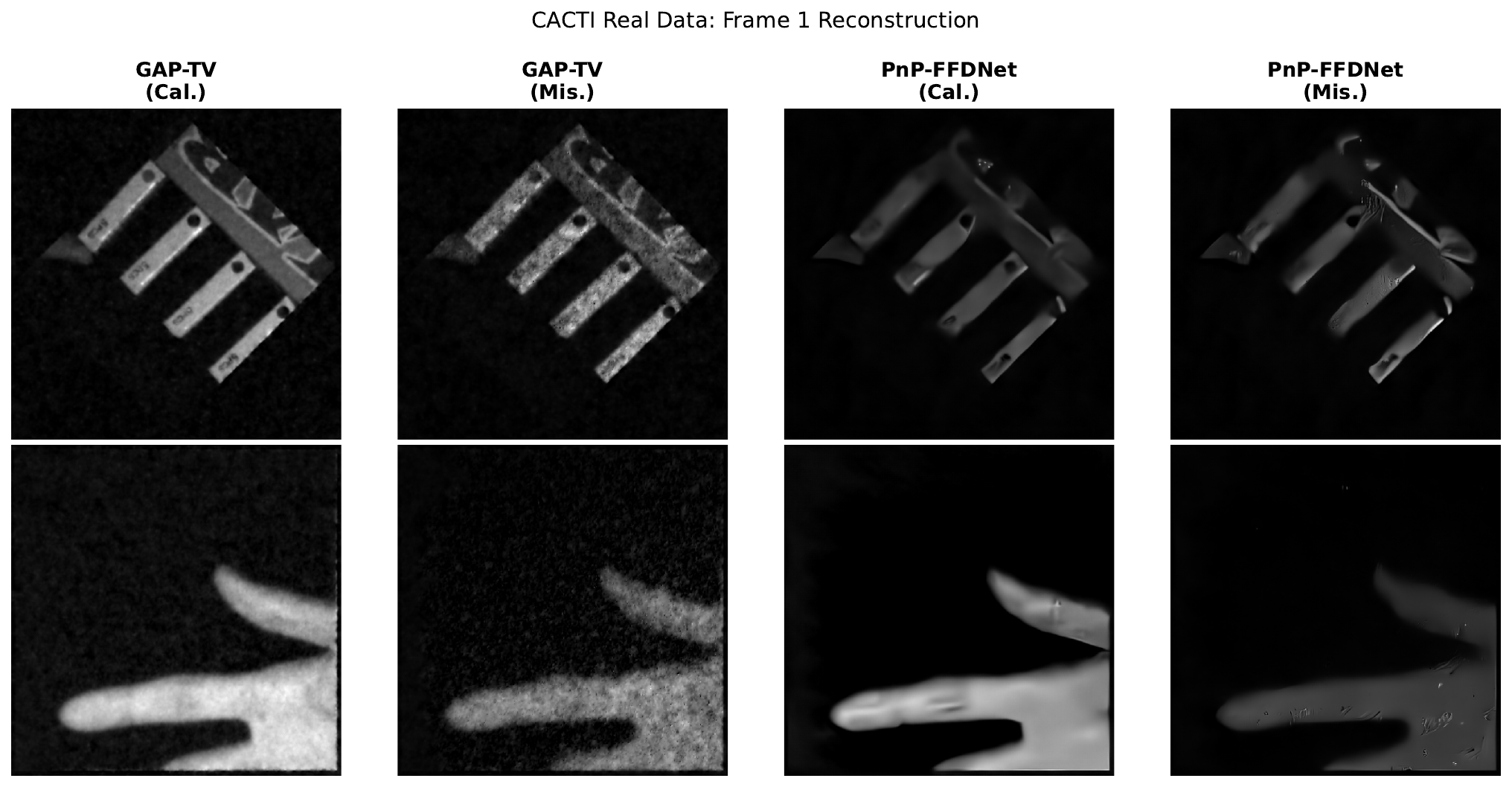}
\caption{CACTI real data: calibrated vs.\ mismatched reconstruction for \textit{duomino} and \textit{hand} scenes. Mismatched masks produce temporal ghosting artifacts, mirroring simulation findings.}
\label{fig:cacti_real}
\end{figure}

\paragraph{CACTI real data.}
\label{sec:cacti_real_data}
We evaluate 4 real CACTI scenes from the EfficientSCI dataset (cr$=$10), using the same measurement residual metric (no ground truth).
\Cref{tab:cacti_real} shows the results.
Under mask mismatch, GAP-TV residuals increase $9.4$--$11.0\times$, confirming mismatch severely degrades real data fidelity.
PnP-FFDNet shows increases of $1.3$--$2.8\times$ (mean $2.0\times$), higher than GAP-TV's baseline but moderated by FFDNet's learned denoiser.
The large difference in absolute calibrated residuals between GAP-TV (1.6e-5) and PnP-FFDNet (0.0041) reflects their different reconstruction mechanisms: GAP-TV minimises the measurement residual directly as its optimisation objective, so the calibrated residual is near-zero by construction; PnP-FFDNet's outer loop is moderated by the denoiser prior, which prevents full data fidelity and results in a higher but still consistent baseline residual.
The \emph{ratio} (mismatched\,/\,calibrated) is therefore the appropriate comparison metric, not the absolute values.
Mismatched reconstructions exhibit temporal ghosting (\cref{fig:cacti_real}).

\begin{table}[t]
\centering
\caption{CACTI real data (4 scenes, $512 \times 512$, cr$=$10). Normalised measurement residual (lower is better). Ratio: mismatched\,/\,calibrated.}
\label{tab:cacti_real}
\begin{tabular}{@{}lccc@{}}
\toprule
\textbf{Method} & \textbf{Calibrated} & \textbf{Mismatched} & \textbf{Ratio} \\
\midrule
GAP-TV     & 1.6e-5 & 1.6e-4 & 10.4$\times$ \\
PnP-FFDNet & 0.0041 & 0.0069 & 2.0$\times$ \\
\bottomrule
\end{tabular}
\end{table}

\subsection{Scenario~IV: Blind Calibration Baseline}
\label{sec:scenario_iv}

Scenario~III assumes oracle knowledge of the true operator.
In practice, the mismatch parameters must be estimated from measurements alone.
Scenario~IV evaluates blind calibration via grid search over mismatch parameters.
For \emph{geometric} mismatch (CASSI, CACTI), the measurement residual provides a strong calibration signal:
\begin{equation}
    \tilde{\boldsymbol{\theta}} = \arg\min_{\boldsymbol{\theta}} \left\| \mathbf{y} - \mathbf{\Phi}(\boldsymbol{\theta})\, \hat{\mathbf{x}}\!\left(\mathbf{y}, \mathbf{\Phi}(\boldsymbol{\theta})\right) \right\|^2,
    \label{eq:scenario_iv}
\end{equation}
where $\boldsymbol{\theta}$ denotes the mismatch parameters and $\hat{\mathbf{x}}(\mathbf{y}, \mathbf{\Phi})$ is the reconstruction from a fast inner-loop solver.
For \emph{radiometric} mismatch (SPC gain drift), the measurement residual is uninformative because the underdetermined system always achieves near-zero self-consistent residual regardless of gain.
Instead, we use reconstruction sparsity (total variation) as the objective:
\begin{equation}
    \tilde{\alpha} = \arg\min_{\alpha} \mathrm{TV}\!\left(\hat{\mathbf{x}}\!\left(\mathbf{y},\, \mathbf{\Phi}(\alpha)\right)\right),
    \label{eq:scenario_iv_tv}
\end{equation}
where $\mathrm{TV}(\cdot)$ denotes the anisotropic total variation.
The rationale is that the correctly calibrated gain produces a clean corrected measurement, yielding a smooth natural-image reconstruction with low TV; incorrect gain leaves systematic artifacts that increase TV.
No ground truth is required for either criterion.

\paragraph{Procedure.}
For each candidate $\boldsymbol{\theta}$, we: (1)~construct the trial operator $\mathbf{\Phi}(\boldsymbol{\theta})$, (2)~reconstruct $\hat{\mathbf{x}}$ using a fast classical solver, and (3)~evaluate the objective (measurement residual for CASSI/CACTI, TV for SPC).
The parameter yielding the lowest objective is selected, and a final high-quality reconstruction is performed with that operator.
Calibration uses a subset of the data: 3 KAIST scenes for CASSI, 2 benchmark videos for CACTI, and 2 Set11 images for SPC---all from the same datasets as \cref{sec:cassi,sec:cacti,sec:spc}, with the same mismatch parameters.

\paragraph{CASSI and CACTI results.}
For CACTI, grid search over mask shifts $dx, dy \in [-1.0, 1.0]$\,px ($9 \times 9$ grid) achieves \emph{near-perfect} calibration: 26.99\,dB (matching Scenario~III), recovering the full 9.39\,dB mismatch gap.
\emph{Important scope caveat}: this grid search calibrates only 2 of the 8 CACTI mismatch parameters ($dx$ and $dy$); the remaining 6 parameters (rotation $\theta$, temporal offset $\Delta t$, duty cycle $\eta$, gain $g$, offset $o$, noise $\sigma_n$) are held at their ground-truth values during calibration. The result therefore measures the upper bound of what 2-parameter spatial calibration can achieve, not full 8-parameter recovery. Extending to the full mismatch space is left to future work.
For CASSI ($11 \times 11$ grid over $dx, dy$), calibration recovers 85\% of the spatial gap ($+1.44$\,dB of $+1.69$\,dB possible); the residual 15\% reflects spatial estimation error ($\hat{dx}{=}0.4$ vs.\ true $0.5$\,px).
The measurement residual provides a strong signal for geometric mismatch because mask shifts cause large, structured data inconsistencies.

\paragraph{SPC results.}
\Cref{tab:spc_scenario_iv} shows the SPC calibration results using TV minimisation (\cref{eq:scenario_iv_tv}).
Grid search over $\alpha \in [0, 0.005]$ with 41 points ($33 \times 33$ blocks, ISTA-Net's learned $\mathbf{\Phi}$) estimates $\hat{\alpha} = 0.00125$ vs.\ true $\alpha = 0.0015$ for both methods, recovering 86--92\% of the oracle bound.
The TV surface exhibits a clear bowl-shaped minimum near the true $\alpha$, confirming that reconstruction sparsity is a viable calibration objective for radiometric mismatch.
PnP-DRUNet achieves higher recovery (92\%) than FISTA-TV (86\%) because the learned denoiser prior produces sharper reconstructions whose TV is more sensitive to residual gain artifacts.
This contrasts with the measurement residual, which is flat across all $\alpha$ values for SPC (the underdetermined system can always self-consistently fit any gain-corrected measurements).

\begin{table}[t]
\centering
\caption{SPC blind calibration via TV minimisation (\cref{eq:scenario_iv_tv}). II: mismatched, no calibration. IV: estimated $\hat{\alpha}$ via TV grid search. III: oracle (true $\alpha$). Recovery: (IV$-$II)\,/\,(III$-$II). True $\alpha{=}0.0015$; estimated $\hat{\alpha}{=}0.00125$.}
\label{tab:spc_scenario_iv}
\begin{tabular}{@{}lcccc@{}}
\toprule
\textbf{Method} & \textbf{II} & \textbf{IV} & \textbf{III} & \textbf{Recovery} \\
\midrule
FISTA-TV   & 19.78 & 26.54 & 27.60 & 86\% \\
PnP-DRUNet & 18.34 & 25.39 & 26.01 & 92\% \\
\bottomrule
\end{tabular}
\end{table}

\paragraph{Criterion comparison.}
The geometric/radiometric distinction reveals that blind calibration requires \emph{matching the objective to the mismatch type}: measurement residual for geometric mismatch (where operator errors create large data infidelities), and reconstruction sparsity for radiometric mismatch (where the operator structure is preserved but measurement values are scaled).
Across all three modalities, Scenario~IV recovers 85--100\% of the oracle bound, demonstrating that blind calibration is practical for all mismatch types studied.

\section{Discussion}
\label{sec:discussion}

\paragraph{Classical vs.\ deep learning robustness.}
Classical methods are consistently more robust: GAP-TV loses 3.38\,dB on CASSI and 10.94\,dB on CACTI, vs.\ 13.98\,dB and 20.58\,dB for the best deep networks.
Under mismatch, the performance hierarchy \emph{inverts}: on CACTI Scenario~II, GAP-TV (15.81\,dB) outperforms EfficientSCI (14.81\,dB) despite being 8.64\,dB worse under ideal conditions---confirming that physical model fidelity dominates algorithmic sophistication.

\paragraph{The operator-awareness spectrum.}
Methods exist on a spectrum: \emph{mask-oblivious} (HDNet, $\rho{=}0\%$) cannot benefit from calibration; \emph{operator-conditioned} (MST, HATNet, $\rho{=}41$--$90\%$) achieve high calibration gains but suffer the largest degradation; \emph{operator-iterative} (GAP-TV, FISTA-TV, $\rho{=}81$--$93\%$ on CACTI/SPC) use the operator directly in each iteration.
This taxonomy reframes the problem: the critical bottleneck is not algorithmic sophistication but \emph{physical model fidelity}.

\paragraph{Practical implications.}
When recalibration is feasible, operator-conditioned networks should be paired with Scenario~IV-style calibration.
When calibration is impractical, classical methods provide the most robust baseline, with degradation 3--5$\times$ smaller than deep methods.
Scenario~IV demonstrates that simple grid-search calibration recovers 85--100\% of the oracle bound without ground truth, provided the objective matches the mismatch type: measurement residual for geometric mismatch, reconstruction sparsity for radiometric mismatch.

\paragraph{Limitations.}
Our parametric mismatch models (affine shifts, gain drift) capture dominant error sources but do not cover spatially varying PSF errors or nonlinear detector response.
Real hardware validation covers CASSI and CACTI; SPC real data would strengthen generalisability.
The grid-search calibration does not scale to high-dimensional spaces---gradient-based or learned calibration is a natural next step.
The benchmark currently evaluates 4 methods per modality; expanding to include recent advances (DAUHST~\cite{cai2022dauhst}, CST~\cite{cai2022cst}, DiffSCI~\cite{meng2024diffsci}) would broaden coverage and further validate the observed trends.
The within-modality inverse performance--robustness relationship is observed over $N{=}4$ methods; cross-modality pooling (12 methods total, Spearman $r_s{=}{-}0.71$) provides stronger statistical support but future work should increase per-modality sample size.

\paragraph{Residual gap analysis.}
The residual gap $\Delta_{\text{res}} = \text{PSNR}_{\text{I}} - \text{PSNR}_{\text{III}}$ measures unrecoverable losses.
CASSI shows the largest residual gaps (MST-L: 7.48\,dB) due to fixed-step dispersion assumptions, while CACTI (GAP-TV: 0.74\,dB) and SPC (HATNet: 1.20\,dB) confirm that spatial and gain-type mismatches are nearly fully recoverable.
Per-method residual gap visualizations are provided in the supplementary material.

\section{Conclusion}
\label{sec:conclusion}

We have presented InverseNet, the first cross-modality benchmark for operator mismatch in compressive imaging.
Evaluating 12 methods across CASSI, CACTI, and SPC under a four-scenario protocol, we find:
(1)~mismatch degrades deep learning methods by 10--21\,dB, collapsing their advantage over classical methods;
(2)~operator-conditioned architectures recover 40--90\% of losses through calibration, while mask-oblivious ones recover 0\%;
(3)~blind grid-search calibration (Scenario~IV) recovers 85--100\% of the oracle bound without ground truth, using measurement residual for geometric and reconstruction sparsity for radiometric mismatch;
(4)~real hardware experiments confirm simulation patterns transfer to physical data.
When calibration is feasible, operator-conditioned networks paired with self-supervised calibration are optimal; otherwise, classical methods provide the most robust baseline.

All reconstruction arrays, metrics, and code will be released upon acceptance.
Future work includes gradient-based calibration, dispersion-aware architectures, and expansion to lensless imaging and ptychography.

\paragraph{Data availability.}
All test data used in this work come from existing public datasets: KAIST~\cite{choi2017kaist} for CASSI, the CACTI benchmark~\cite{yuan2021snapshot}, and Set11~\cite{kulkarni2016reconnet} for SPC. Real CASSI scenes are from the TSA dataset~\cite{wagadarikar2008cassi}; real CACTI scenes are from the EfficientSCI repository~\cite{wang2023efficientsci}. No non-public datasets were used. Code and reconstruction arrays are available at \url{https://github.com/integritynoble/Physics_World_Model}.

\bibliographystyle{splncs04}

\end{document}